\def\year{2018}\relax
\documentclass[letterpaper]{article} 
\usepackage{aaai18}  
\usepackage{times}  
\usepackage{helvet}  
\usepackage{courier}  
\usepackage{url}  
\usepackage{graphicx}  

\usepackage{amsmath}
\usepackage{CJKutf8}
\usepackage{algorithm}
\usepackage[noend]{algpseudocode}
\usepackage{multirow}
\usepackage{colortbl}

\newcommand{\citet}[1]{\citeauthor{#1}~(\citeyear{#1})}
\begin{document}
%
\title{Image Inspired Poetry Generation in XiaoIce\thanks{The work was done when the first author and the second author worked as interns in Microsoft.}}

\author{Wen-Feng Cheng$^{1,2}$, Chao-Chung Wu$^2$, Ruihua Song$^1$, Jianlong Fu$^1$, Xing Xie$^1$, Jian-Yun Nie$^3$\\
$^1$Microsoft, $^2$National Taiwan University, $^3$University of Montreal\\
\{wencheng, rsong, jianf, xingx\}@microsoft.com, r05922042@ntu.edu.tw, nie@iro.umontreal.ca\\
}

\maketitle

\begin{abstract}
Vision is a common source of inspiration for poetry. The objects and the sentimental imprints that one perceives from an image may lead to various feelings depending on the reader. In this paper, we present a system of poetry generation from images to mimic the process. Given an image, we first extract a few keywords representing objects and sentiments perceived from the image. These keywords are then expanded to related ones based on their associations in human written poems. Finally, verses are generated gradually from the keywords using recurrent neural networks trained on existing poems. Our approach is evaluated by human assessors and compared to other generation baselines. The results show that our method can generate poems that are more artistic than the baseline methods. This is one of the few attempts to generate poetry from images. 
By deploying our proposed approach, XiaoIce\footnote{XiaoIce is a Microsoft AI product popular on various social platforms, focusing on emotional engagement and content creation\cite{xiaoice}} has already generated more than 12 million poems for users since its release in July 2017. A book of its poems has been published by Cheers Publishing, which claimed that the book is the first-ever poetry collection written by an AI in human history.
\end{abstract}

\section{Introduction}
\begin{figure*}[!ht]
	\begin{center}
		\includegraphics*[width=0.95\textwidth]{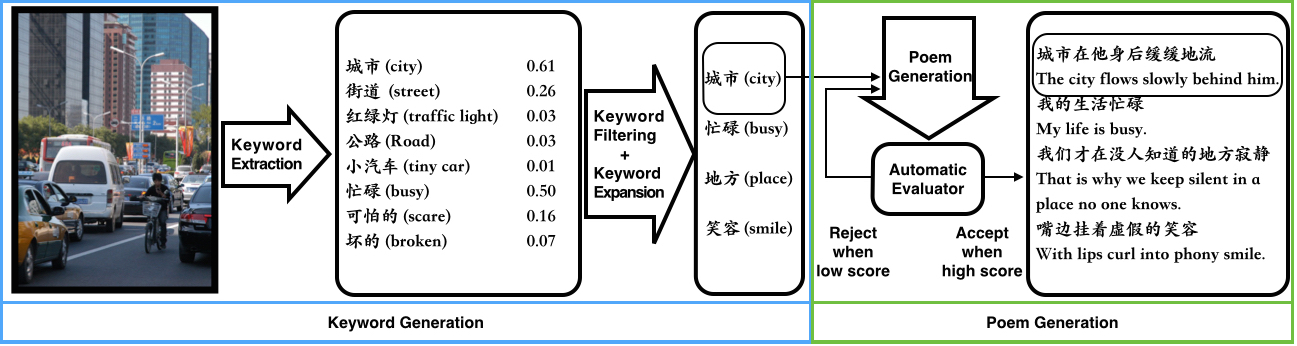}
		\caption{Illustration of the Image to Poetry framework. The system takes an image query given by a user, and outputs a semantically relevant piece of modern Chinese poetry. For the left part of the figure, after the intermediate keywords extracted from the query by object and sentiment detection, keyword filtering and expansion are applied to generate a keyword set. After that, each keyword in the keyword set is considered as a seed for each line in the poem, as shown in the poem generation part. A hierarchical generating model is proposed to maintain both the fluency of sentences and the coherence between sentences. In addition, an automatic evaluator is used to select sentences with good quality.            
			\label{fig:Fig1}
		}	
	\end{center}
\end{figure*}

Poetry is always important and fascinating in Chinese literature, not only in traditional Chinese poetry but also in modern Chinese poetry. 
While traditional Chinese poetry is constructed with  strict rules and patterns (e.g., five-word quatrains are required to contain four sentences and each sentence has five Chinese characters, also words need rhymes in specific positions), modern Chinese poetry is unstructured in vernacular Chinese. Compared to traditional Chinese poetry, although the readability of vernacular Chinese makes modern Chinese poetry easier to strike a chord, errors in words or grammar can more easily be criticized by users. Good modern poetry also requires more imagination and creative uses of language. From these perspectives, it may be more difficult to generate a good modern poem than a classic poem. 

Poetry can be inspired by many things, among which vision (and images) is certainly a major source. Indeed, poetic feelings may emerge when one contemplates an image (which may represent anything from a natural scene to a painting). It is usually the case that different people have different readings and feelings of the same image. This makes it particularly interesting to read poems by others inspired by the same image. In this work, we present a system that mimic poetry writing of a poet by looking at an image. 

Generating poetry from image is a special task of text generation from image. There have been many studies in this area.
However, most of them focus on image captioning rather than literature creation. Only few of previous systems addressed the problem of generating poems from images. There have also many studies and systems for generating poetry. In most cases, a system is provided with a few keywords and is required to compose a poem containing or relating to the keywords. In comparison with poetry generation from keywords, using image as inspiration for poetry has many advantages. First, an image is worth thousand words, and it contains richer information than keywords. Poems generated from images maybe more various. Second, as mentioned earlier, for different people, the same image could lead to different interpretations, thus using images to inspire poetry generation may often provide an enjoyable surprise and leave the impression of higher imagination. Finally, compared with asking users for providing keywords, uploading an image is a much simpler and more nature way to interact with a system nowadays.

The system we propose, as illustrated in Figure~\ref{fig:Fig1}, aims to generate a modern Chinese poem inspired by a visual content. For the image on the left hand side, we extract objects and sentiments to form our initial keyword set, such as \textit{city} and \textit{busy}. Then, the keywords are filtered and expanded by associated objects and feelings.
Finally, each keyword is regarded as an initial seed for each sentence in the poem. A hierarchical recurrent neural network is used for modeling the structure between words and between sentences, and a fluency checker automatically detects low quality sentences early so that a new sentence is generated when necessary.   

Our main contributions are as follows:
\begin{itemize}
	\item We introduce a novel application that uses an image to inspire modern poetry generation, which mimic the human behaviors of expressing their feelings when they are touched by vision.  
	\item 
	In order to generate poetry of good quality, we incorporate several verification mechanisms for text fluency, poetry integrity, and the matching with the image. 
	\item 
	We leverage keyword expansion to improve the diversity of generated poems and make them more imaginative.
\end{itemize}

A book of 139 generated poems, titled ``Sunshine Misses Windows'', was published on May 19, 2017 by Cheers Publishing, which claimed that the book is the first-ever poetry collection written by an AI in human history. We also release the system in XiaoIce products in July, 2017. As by August, 2018, about 12 million poems have been generated for users.

The rest of the paper is organized as follows. Section~\ref{sec:rw} includes related works on image caption and poetry generation. Section~\ref{sec:ip} describes the details of the problem and our approach. The training details are explained in Section~\ref{sec:td} and the datasets and experiments are presented in Section~\ref{sec:eoa}. We also design a user study to compare our approach with state-of-the-art image to caption and CTRIP (the only known poetry generation system from image) in Section~\ref{sec:ec}. Section~\ref{sec:cf} concludes this paper.

\section{Related Work}
\label{sec:rw}
Image to caption has been a popular research topic in recent years. \cite{bernardi2016automatic} provides an overview of most image description research and classifies approaches into three categories. Our work would be categorized as ``Description as Generation from Visual Input'', which takes visual features or information from images as input for text generation. \cite{patterson2014sun} and \cite{devlin2015language} regard descriptions as retrieved results in the visual space. Although they can retrieve grammatically correct sentences and be applicable to novel images, the quality greatly depends on the training dataset. Among the works similar to ours which exploit visual input to description generation, RNN-based models achieve great quality recently. \cite{Socher2014GroundedCS} maps image and sentence representation to a latent space so that text and image become related. \cite{DBLP:journals/aimatters/SotoKKM15} exploits a decoder-encoder framework. \cite{KarpathyL15} and \cite{donahue2015long} apply either LSTM architecture or alignment of image and sentence models for further improvement. However, most of them need image-sentence pairs for training. For image to poetry, there is no existing large scale data of paired images and poems.

Along with the glorious poetry history, automatic poetry generation is another popular research topic in artificial intelligence, starting from the Stochastische Texte system (Lutz 1959). Like the system, the first few generators are template-based. \cite{tosa2008hitch} and \cite{wu2009new} developed an interactive system for traditional Japanese Poetry. \cite{oliveira2012poetryme} proposed a system based on semantic and grammar templates. Word association rules were applied in \cite{netzer2009gaiku}. The systems based on templates and rules can generate sentences with have correct grammar but this is at the price of less flexibility. As the second type of generator, genetic algorithms are applied in previous works, like \cite{manurung2004evolutionary} and \cite{manurung2012using}, which regard poetry generation as a state search. \cite{yan2013poet} formulate the task as an optimization problem based on a generative summarization framework under several constraints. \cite{jiang2008generating} present a phrase-based statistical machine translation to generate the second sentence from the first sentence. \cite{he2012generating} extend the approach to a sequential translation for quatrains.

The growth of deep learning also brings success to poem generation. The basic recurrent neural network language model (RNNLM) \cite{mikolov2010recurrent} can generate poetry by using poetry corpus. \cite{zhang2014chinese} generated lines incrementally instead of regarding a poem as a single sequence. \cite{yan2016poet} added an iterative polishing to a hierarchical architecture. \cite{wang2016chinese} applied the attention-based model. \cite{yi2016generating} extended the approach into a quatrain generator with an input word as a topic. \cite{ghazvininejad2016generating} generated poems on a user-supplied topic with rhythmic and rhyme constraints. \cite{wang2016chinese2} proposed planning-based method to ensure the poem coherence and consistency. All these studies focus on the problem of generating a poem from a text input. None of them involves non-textual modality.

There have been other studies connecting multiple modalities. \cite{schwarz2016auto} connected images and poetry by automatically illustrating poems via semantically relevant and visually coherent illustrations. However, the task is not to generate a poem from an image, which is a more complex task. Our work, focuses on automatically generating a semantically relevant poem from an image.
\begin{figure*}[!ht]
	\begin{center}
		\includegraphics[width=0.7\textwidth]{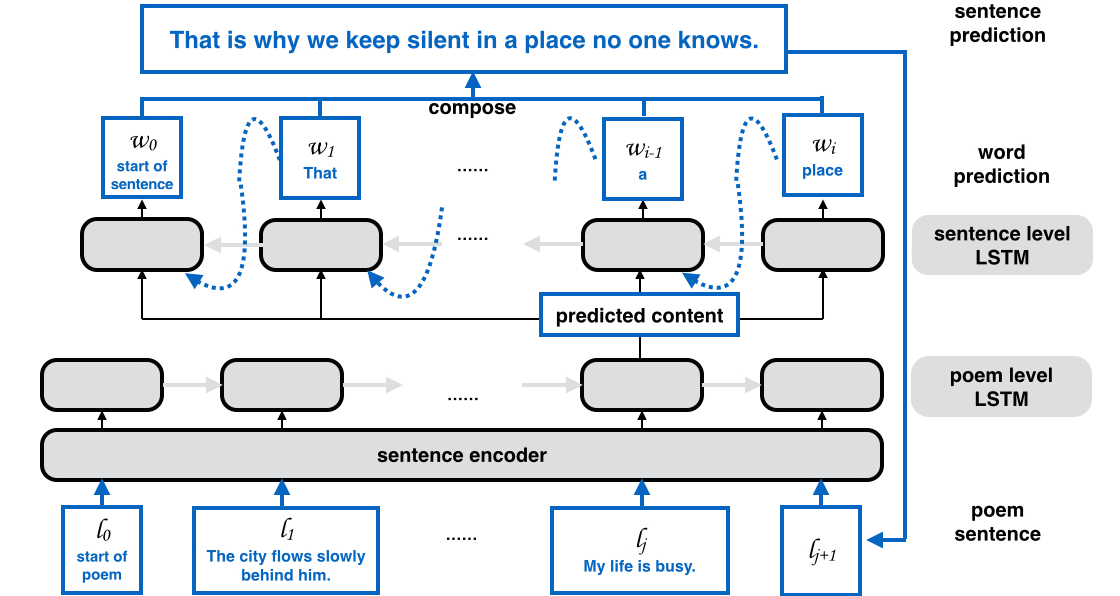}
		\caption{Our proposed hierarchical poem model includes two levels of LSTM. With the poem level model illustrated in the lower half of the figure, we predict the content vector of the next sentence by considering all previous sentences. After that, the content vector will be regarded as an input of sentence level LSTM in the upper half of the figure. Notice that this figure only shows the backward generator for recursive generating, while the forward version can be modified by reversing the structure.                  
			\label{fig:model}
		}	
	\end{center}
\end{figure*} 

\section{Image to Poetry}
\label{sec:ip}
\subsection{Problem Formulation and System Overview}

To achieve the goal of generating poems inspired by image, we formulate the problem as follows: for an image query $\boldsymbol{Q}$, we try to generate a poem $\boldsymbol{P}=({l}_1, {l}_2, \cdots, {l}_N)$, where ${{l}_i}$ represents the ${i}$-th line of the poem and ${N}$ is the number of lines in poem. The poem is supposed to be relevant to the image content, fluent in language and coherent in semantics.

The overview of our solution is shown in Figure~\ref{fig:Fig1}. For the image query, object and sentiment detection are used to extract appropriate nouns, such as \textit{city} and \textit{street}, and adjectives, such as \textit{busy}, as initial keyword set. After filtering out words with low confidence and rare words, keyword expansion will be applied to construct a keyword set $\boldsymbol{K}=({k}_1,{k}_2,\cdots,{k}_N)$, whose size is equal to lines of the poem. In the example, \textit{place} and \textit{smile} are expanded. Now $\boldsymbol{K}$ contains four keywords, i.e. \textit{city}, \textit{busy}, \textit{place} and \textit{smile}. Next, each keyword is regarded as an initial seed for each sentence in the poem generation process. For example, the first sentence is generated from the seed \textit{city}. A hierarchical recurrent neural network is proposed for modifying the structure between words and between sentences. Finally we apply a fluency checker to automatically detect low quality sentences early on and re-generate them.   

We use Long-Short Term Memory (LSTM) for RNN mentioned below. The basic element for generation could be a character or a word. We try both in our experiments. 

\subsection{Keyword Extraction}
We propose detecting objects and sentiments from each image with two parallel convolutional neural networks (CNN), which share the same network architecture but with different parameters. Specifically, one network learns to describe objects by the output of noun words, and the other learns to understand the sentiments by the output of adjective words. The two CNNs are pre-trained on ImageNet \cite{imagenet} and fine-tuned on noun and adjective categories, respectively. For each CNN, the extracted deep convolutional representations are denoted as $\boldsymbol{{W}_c}\ast\boldsymbol{I}$, where $\boldsymbol{{W}_c}$ denotes the overall parameters of one CNN, $\ast$ denotes a set of operations of convolution, pooling and activation, and $\boldsymbol{I}$ denotes the input image. Based on deep representation, we further generate a probability distribution ${P}$ over the output object or sentiment categories $\boldsymbol{C}$, shown as:
\begin{eqnarray}\label{prob}
	P(\boldsymbol{C}\mid\boldsymbol{I}) = f(\boldsymbol{{W}_c} \ast\boldsymbol{I}),
\end{eqnarray}
where $f(\cdot)$ represents fully-connected layers to map convolutional features to a feature vector that could be matched with the category entries, and includes a softmax layer to further transform the feature vector to probabilities. For the proposed parallel CNN, we denote the probability over noun and adjective categories as $p^{n}(\boldsymbol{C}\mid\boldsymbol{I})$ and $p^{a}(\boldsymbol{C}\mid\boldsymbol{I})$, respectively. Categories with high probabilities are chosen to construct the candidate keyword set.
\subsection{Sentence Model}
\textbf{RNNLM }
We follow the recurrent neural network language model (RNNLM) \cite{mikolov2010recurrent} to predict text sequence. Each word is predicted sequentially by the previous word sequence: 
\begin{eqnarray}\label{f_prob}
	w_i = \arg\max_{w}P(w\mid w_{1:i-1}),
\end{eqnarray}
where $w_i$ is the $i$-th word and $w_{1:i-1}$ means the preceding words sequence. \\
\textbf{Recursive Generation }
To control the content of generated sentences, we use specific keywords as the seed for sentence generation, which means that we force the RNNLM to generate sentence with specific keywords. Due to the directivity of RNNLM, one can only generate forward from the existing word. To allow the keyword to appear at any position in a sentence, a simple idea is training a reverse version of RNNLMs (which input the corpus by a reverse ordering in training), and generating backward from the existing text: 
\begin{eqnarray}\label{b_prob}
	w_i = \arg\max_{w}P(w\mid w_{n},w_{n-1},...,w_{i+1}). 
\end{eqnarray}
However, if we generate the forward and backward separately, the result would be two independent parts without semantic connections. To solve this problem, we use a simple recursive strategy described below.

Let $\mathbf{<sos>}$ and $\mathbf{<eos>}$ represent the start symbol and end symbol of a sentence. Also, $\mathbf{LM_{forward}}$ and $\mathbf{LM_{backward}}$ are the original and reversed version of RNNLM. The process of generating the $j$-th line $l_j$ with $j$-th keyword $k_j$ in the poem is described in Algorithm~\ref{recur}.
\begin{algorithm}
	\caption{Recursive Generator\label{recur}}
	\begin{algorithmic}[1]
		
		\State $\textit{sequence} \gets {k_j}$
		\While{$\mathbf{<sos>}\notin\textit{sequence}$  
			\textbf{or}
			$\mathbf{<eos>}\notin\textit{sequence}$}
		\If {$\mathbf{<eos>}\notin\textit{sequence}$}
		\State $w=\arg\max_{w}P(w\mid\textit{sequence},\mathbf{LM_{forward}})$ 
		\State $\textit{sequence} \gets \textit{sequence}+w$
		\EndIf
		\If {$\mathbf{<sos>}\notin\textit{sequence}$}
		\State 
		$w=\arg\max_{w}P(w\mid\textit{sequence},\mathbf{LM_{backward}})$ 
		\State $\textit{sequence} \gets w+\textit{sequence}$
		\EndIf
		\EndWhile
		\State $l_j \gets \textit{sequence}$
	\end{algorithmic}
\end{algorithm}
\subsection{Poem Model}
\textbf{Generation with Previous Line }
While the fluency of sentences can be controlled with the RNNLM model and a recursive strategy, in multi-keyword and multi-line scenario, another issue is to maintain consistency between sentences. Since we need to generate in two directions, using the state of RNNLM to pass the information is no longer feasible. Instead, we try to extend the input gate of the RNNLM model to two parts, one is the originally previous word input, and another is the previous sentence's information. Here, we use the encoding of previous line by LSTM as input context. For generating ${j}$-th line ${l_j}$ in the poem, we use:
\begin{equation}\label{preline}
	w_i = \arg\max_{w}P(w\mid w_{1:i-1},l_{j-1}).
\end{equation}
\textbf{Hierarchical Poem Model }
Although the model above can maintain the consistency of a poem by capturing the previous line's information, an alternative idea is to maintain a poem level network. For the poem level, we try to predict the content vector of the next sentence by all previous sentences, and for the sentence level, we use the prediction as another input. By using the hierarchical structure as shown in Figure~\ref{fig:model}, we can maintain the fluency and consistency not only using the previous line but also all previous lines. For generating ${j}$-th line ${l_j}$ in the poem, we use:
\begin{equation}\label{linelstm}
	w_i = \arg\max_{w}P(w\mid w_{1:i-1},l_{1:j-1}).
\end{equation}
Notice that since we still need to use the recursive strategy described above, a forward version and a backward version of models are both required.

\subsection{Keyword Expansion}
Since we attempt to control the generation using the objects and sentiments as keywords, the final results would correspond closely to the keywords. However, two possible reasons might lead to failure: low confidence keywords and rare keywords. The former is caused by the limitation of the image recognition model which will make the generated sentences irrelevant to the query images, while the latter will lead to low-quality or monotonous generated sentences due to insufficient training data. Thus we choose to use the keywords that have not only high confidence in image recognition but also enough occurrences in training corpus. However, sometimes the number of image keywords may be less than $N$. 
Even if the number of initial keywords is larger than N, keyword expansion is also useful, it allows us to go beyond what is directly observable from the image. Using such expanded keywords could make the poetry more imaginative and less descriptive. In this work, we test several options:\\
\textbf{Without Expansion }
The first idea is simple, we can choose keywords only with high confidence in recognition and enough occurrences in corpus. While these keywords can be considered seeds for the recursive generation with forward and backward models, for the rest of the poem, without giving any new keywords, we only generate new lines according to the previous line by the forward model only.\\
\textbf{Frequent Words }
To expand the keyword set, an approach is to select some frequent nouns and adjectives in the training corpus. After deleting the rare and inappropriate words, the expanded keywords are sampled with the word distribution of the training corpus. The higher frequency a word is, the greater the chance it will get in. The three nouns with the highest frequency of occurrence in our corpus are \textit{life}, \textit{time}, and \textit{place}. Applying these words can enhance both the diversity and imagination of the generated poems without getting off the topic too much.\\
\textbf{High Co-occurred Words }
Another idea is only considering words with high co-occurrence with the original image keywords in the training corpus. We sample the keywords with the distribution of co-occurrence frequency with the original keywords. The more often a word co-occurs with the selected image keyword, the greater the chance it will get in. Take \textit{city} for example, words with highest co-occurrence with \textit{city} are \textit{place}, \textit{child}, \textit{heart} and \textit{land}. Unlike the previous method, these words are usually more relevant to the keywords recognized from the image query, hence the result is expected to be more on topic.
\subsection{Fluency Evaluator}
In poetry, it is desirable to generate diverse sentences even for the same keyword, we randomly sample word candidates among the top n best in beam search. This resulting sentence may be the one never seen in training data. At the same time, we can generate diverse sentences for images with same objects or sentiments. However, this diversity in the generation process may lead to poor sentences which are not fluent or inconsistent semantically.

To overcome these issues, we use an automatic evaluator of a sentence. We use n-gram and skip n-gram models to measure whether a word is correct and whether two words have semantic consistency. For the grammar level, we train a LSTM-based language model with POS tagged corpus and then apply it to calculate the generation probabilities of POS tagged candidate sentences. The failure to pass the evaluation will lead to the generation of another sentence.

\section{Training Details}
\label{sec:td}

As a training corpus, we collect 2,027 modern Chinese poems that are composed of 45,729 sentences from shigeku.org. The character vocabulary size is 4,547. For the training of word based model, word segmentation are applied on the corpus. The size of word vocabulary is 54,318.

In the keyword extraction model, for each CNN in our parallel architecture, we select GoogleNet \cite{googleNet} as the basic network structure, since GoogleNet can produce the first-class performance on ImageNet competition \cite{imagenet}. Following \cite{TaggingTransferDeep} and \cite{VSO}, we use 272 nouns and 181 adjectives as the categories for noun and adjective CNN training, since these categories are adequate to describe common objects and sentiments conveyed from images. The training time for each CNN takes about 50 hours on a Tesla K40 GPU, with the top-1 classification accuracy of $92.3\%$ and $85.4\%$ on noun and adjective testing sets from \cite{TaggingTransferDeep} and \cite{VSO}, respectively.

In the poetry generation model, the recurrent hidden layers for the sentence level and poem level both contain 3 layers and 1024 hidden units for each layer. The sentence encoder dimensionality is 64. The model was trained with the Adam optimizer \cite{kingma2014adam}, where 128 is used as the minibatch size. The training time for each CNN takes about 100 hours on a Tesla K80 GPU.

\section{Experiments of Our Approach}
\label{sec:eoa}

The system involves several components with several choices. Since it is hard to measure all combinations as they may influence each other, to optimize our system, we design the experiment process with a greedy strategy and separate the experiment into two parts. For each part, we compare the different method choices for one more step combined with the best approach of the previous experiment. The former is poem generation considering sentence level and poem level models. Since we use keywords as the seed for generation, for the latter, we focus on the quality of keywords from keyword extraction and different keyword expansion methods.
\begin{figure*}[!t]
	\centering
	\includegraphics[width=0.8\pdfpagewidth]{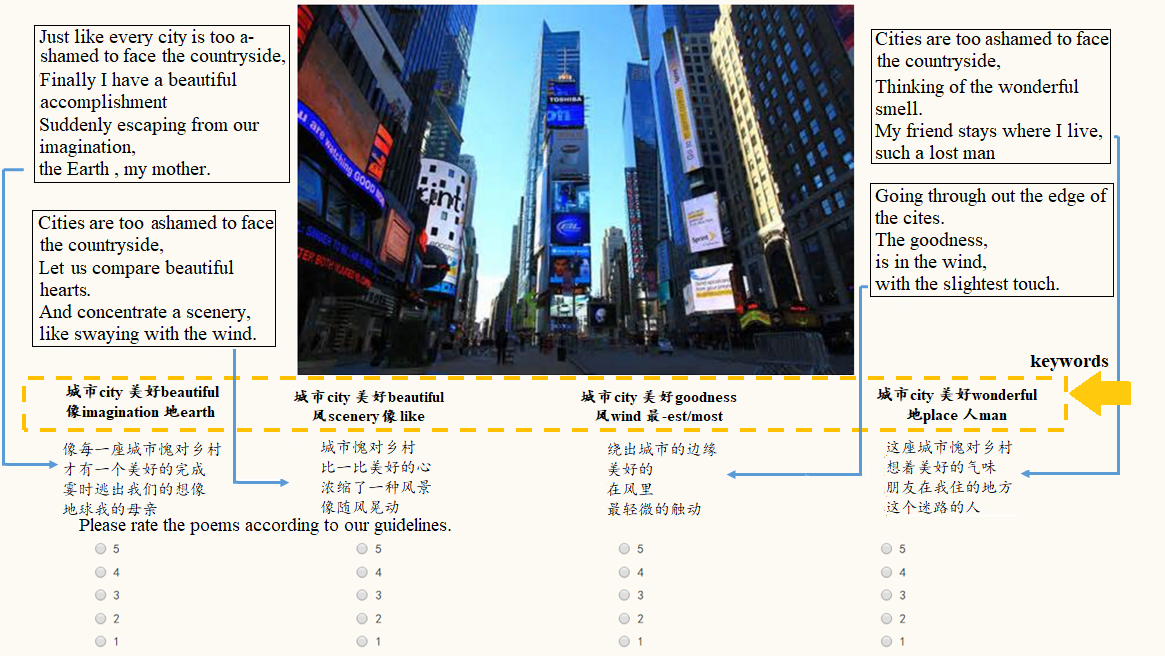}
	\caption{The human assessment tool is designed to capture the relative judgments among methods. For each image, a four-line poem is generated with each method, and all poems are displayed side by side. 
		\label{fig:judge}
	}	
\end{figure*} 
\subsection{Experiment Setup}
\textbf{Test Image Data }
For the model optimization experiment, 100 public domain images are crawled from Bing image search by searching 60 randomly sampled nouns and adjectives in our predefined categories. We focus on 45 images recognized as views for optimizing our model. The data will be released to research communities. Please note although our experiments are conducted on some type of images, our proposed method is general. Actually since we released our system in July, 2017, users have submitted about 1.2 million images of all kinds and gotten created poems by August, 2018.\\
\textbf{Human Evaluation }
As shown in \cite{liu2016not}, overlap-based evaluation metrics, such as BLEU and METEOR, have little correlation with real human feelings. Thus, we conduct user studies to evaluate our method. 

The interface of the judgments is shown in Figure~\ref{fig:judge}. We present an image at the top and poems generated by different methods for comparison side by side below the image. For each poem, we ask assessors to give a rating from 1 (dislike) to 5 (like) after they compare all the poems. We do not choose the design that shows a poem each time and asks for a rating from assessors because such kind of rating is not stable for comparing the quality of poems. The assessors may change their standards unconsciously. Our design borrows the idea of A-B test widely used in search evaluation. When an assessor can easily read and compare all poems before rating, his/her scores can provide meaningful information on relative ordering of the poems. Therefore, we focus on the relative performance in each experiment rather than absolute scores.
In addition, We randomly arrange the order of methods for each image to remove biases of ordering and about a particular method. Due to the high cost of human evaluation, we invite five engineer background college students and two literature background students to judge all methods for optimizing models.
\subsection{Poem generation}
In poem generation, we consider sentence level experiment first. After the best approach is chosen, information from the previous sentence is used in poem level models. \\
\textbf{Sentence Level }
For the sentence level, we aim to figure out whether the recursive generating strategy can produce more fluent sentences with specific keywords (here two nouns and two adjectives). As a baseline, we generate the part before a keyword by a backward model and the part after it by a forward model separately and then combine them. We also consider the influence of using different generating elements (character or word). There are four methods: \textit{char\_combine}, \textit{char\_recursive}, \textit{word\_combine}, and \textit{word\_recursive}.
Although the sentence level experiment focuses on the sentence generation, for the convenience of the users' judgments, we still present a four-line poem with four fixed keywords for each method. As shown in Table~\ref{ex1}, \textit{word\_recursive} is significantly better than the two character-based methods. The \textit{char\_recursive} method is also significantly better than \textit{char\_combine}. Although the difference between two word-based methods is not significant, the \textit{word\_recursive} method is better than the \textit{word\_combine} method, in particular when we compare them with the method \textit{char\_recursive}. The \textit{word\_recursive} method significantly outperforms \textit{char\_recursive} method, while the \textit{word\_combine} does not. Therefore, we choose the \textit{word\_recursive} as the best method in this step.\\
\textbf{Poem Level }
As the best method in the previous level, \textit{word\_recursive} is kept with adding additional methods in the second user study. While \textit{word\_recursive} ignores the relationship between sentences, \textit{word\_preline} uses the encoding of previous lines as an additional input, and \textit{word\_poemlstm} considers all previous lines by maintaining a poem level LSTM. Again, four fixed keywords are given for each method. The results in Table~\ref{ex1} show that both methods bring significant improvements over the \textit{word\_recursive} method. By considering all previous lines, the \textit{word\_poemlstm} method also significantly outperforms the \textit{word\_preline} by about 11\%. This indicates that our proposed hierarchical poem model works the best.
\begin{table}[!t]
	\centering
	\caption{Human evaluation results of our poem generator on sentence level and poem level. The average scores show that both the recursive strategy and hierarchical model gain improvement significantly (with p-value less than 0.01). }
	\label{ex1}
	\resizebox{\columnwidth}{!}{
		\begin{tabular}{|c|c|c|l|l|}
			\hline
			\multicolumn{5}{|c|}{Sentence Level} \\ \hline
			& Average & \multicolumn{3}{c|}{p-value} \\ \cline{3-5} 
			\multirow{-2}{*}{Approaches} & Score & \multicolumn{1}{c|}{\textbf{char\_recursive}} & \multicolumn{1}{c|}{\textbf{word\_combine}} & \textbf{word\_recursive} \\ \hline
			\textbf{char\_combine} & 2.30 & $1.22\times 10^{-3}$ & $9.94\times 10^{-4}$ & $3.52\times 10^{-7}$ \\ \hline
			\textbf{char\_recursive} & 2.55 &  & $1.47\times 10^{-1}$ & $3.58\times 10^{-3}$ \\ \hline
			\textbf{word\_combine} & 2.68 &  &  & $7.95\times 10^{-2}$ \\ \hline
			\textbf{word\_recursive} & {\underline{ \textbf{2.86}}} &   &  &  \\ \hline
			\hline
			\multicolumn{5}{|c|}{Poem Level} \\ \hline
			& Average & \multicolumn{3}{c|}{p-value} \\ \cline{3-5} 
			\multirow{-2}{*}{Approach} & Score & \multicolumn{1}{c|}{\textbf{word\_preline}} & \multicolumn{1}{c|}{\textbf{word\_poemlstm}} &  \\ \hline
			\textbf{word\_recursive} & 2.64 & $5.83\times 10^{-3}$ & $1.76\times 10^{-8}$ &  \\ \hline
			\textbf{word\_preline} & 2.95 &  & $5.14\times 10^{-3}$ &  \\ \hline
			\textbf{word\_poemlstm} & {\underline{ \textbf{3.27}}} &  &  &  \\ \hline
	\end{tabular}}
\end{table}

\subsection{Keyword Extraction and Expansion}
Given the superiority of the \textit{word\_poemlstm} method, we then measure keyword filtering and expansion with this poem generation method. The original four keywords we provide are two nouns and two adjectives with the highest probabilities from the keyword extracting step. We compare the baseline with different keyword expansion methods. The \textit{without\_expand} method only uses two appropriate extracted keywords (one noun and one adjective) after keyword filtering. The \textit{expand\_freq} enlarges the two keywords according to the word frequency in the whole corpus. And the \textit{expand\_cona} expands the two keywords based on the co-occurred frequency with them. While this step focuses on the relevance of keyword to image, besides a 1-to-5 score to each poem, the assessors are also asked to label a true/false label on each keyword and corresponding sentence according to their relevance to the image query.

While there is no obvious difference between the average scores in Table~\ref{ex2}, these four approaches show a totally different performance on the relevance to image query. Since \textit{without\_expand} only uses high confidence keywords, it has the lowest keyword irrelevance rate. However, as the longer generation is to be more problematic, the irrelevance rate of generated sentences by \textit{without\_expand} increases dramatically to 30\%. \textit{expand\_freq} and \textit{expand\_cona}, by contrast, reduce the sentence irrelevance rate by enlarging the keyword set with additional words. While the keyword irrelevance rate of \textit{expand\_freq} increases slightly, by considering the word co-occurrence, \textit{expand\_cona} gains the 
best sentence relevance rate and decreases the keyword irrelevance rate from 18.7\% to 15.6\%.
\begin{table}[!t]
	\centering
	\caption{The performance of different keyword expansion methods. While they share close average scores, both keyword and sentence irrelevance rate drop by applying query expansion based on word co-occurrence. }
	\label{ex2}
	\resizebox{\columnwidth}{!}{
		\begin{tabular}{|c|c|c|c|}
			\hline
			\multirow{2}{*}{Approach} & Average & Keyword & Sentence \\
			& Score & irrelevance rate & irrelevance rate \\ \hline
			\textbf{word\_poemlstm} & 3.10 & 18.7\% & 23.5\% \\ \hline
			\textbf{without\_expand} & 3.11 & {\underline {\textbf{6.9\%}}} & 30.0\% \\ \hline
			\textbf{expand\_freq} & {\underline {\textbf{3.12}}} & 19.4\% & 22.8\% \\ \hline
			\textbf{expand\_cona} & 3.11 & 15.6\% & {\underline {\textbf{21.1\%}}} \\ \hline
		\end{tabular}
	}
\end{table}

\begin{table*}[!t]
	\centering
	\caption{Human evaluation results of our method and two other baselines. While the ``related'' part is dominated by Image2caption and CTRIP is stronger in the ``fluent'' part, our method significantly outperforms both baselines in other aspects.}
	\label{ex4}
	\resizebox{\textwidth}{!}{
		\begin{tabular}{|l|l|l|l|l|l|l|l|l|l|}
			\hline
			
			&  \multicolumn{3}{l|}{\cellcolor[gray]{0.9} Overall} & \multicolumn{3}{l|}{Relevant} & \multicolumn{3}{l|}{Fluent} \\ \hline
			\multirow{2}{*}{Method} & Average & \multicolumn{2}{l|}{p-value} & Positive & \multicolumn{2}{l|}{p-value} & Positive & \multicolumn{2}{l|}{P-value} \\ \cline{3-4} \cline{6-7} \cline{9-10} 
			& score & \textbf{CTRIP}& \textbf{Image2caption} & rate & \textbf{CTRIP} & \textbf{Image2caption} & rate & \textbf{CTRIP} &\textbf{ Image2caption} \\ \hline
			\textbf{our method }&{\underline {\textbf{4.27}}}  &  $2.92\times 10^{-1}$&  $2.67\times 10^{-96}$ &6.3\%  &  $1.8\times 10^{-1}$ &  $0$   & 26.5\%&  $1.61\times 10^{-95}$ &  $2.56\times 10^{-58}$ \\ \hline
			\textbf{CTRIP} & 4.25 &  &  $2.09\times 10^{-97}$ &  5.9\%&  & $0$ &{\underline {\textbf{63.0\%}}}  &  & $5.43\times 10^{-125}$  \\ \hline
			\textbf{Image2caption} & 3.67 &  &  &{\underline {\textbf{82.1\%}}}  &  &  &6.2\%  &  &  \\ \hline
			\hline
			
			& \multicolumn{3}{l|}{Imaginative} & \multicolumn{3}{l|}{Touching} & \multicolumn{3}{l|}{Impressive} \\ \hline
			\multirow{2}{*}{Method} & Positive & \multicolumn{2}{l|}{p-value} & Positive & \multicolumn{2}{l|}{p-value} & Positive & \multicolumn{2}{l|}{p-value} \\ \cline{3-4} \cline{6-7} \cline{9-10} 
			& rate & \textbf{CTRIP} & \textbf{Image2caption} & rate  & \textbf{CTRIP} &\textbf{ Image2caption} & rate  & \textbf{CTRIP} & \textbf{Image2caption} \\ \hline
			\textbf{our method} & {\underline {\textbf{57.0\%}}} & $8.27\times 10^{-142}$   &  $3.70\times 10^{-231}$  & {\underline {\textbf{45.4\%}}} &  $5.11\times 10^{-23}$  &  $5.34\times 10^{-166}$  &{\underline {\textbf{43.4\%}}}\% & $3.67\times 10^{-13}$   & $1.19\times 10^{-152}$   \\ \hline
			\textbf{CTRIP }& 27.7\%&  &  $2.51\times 10^{179}$  &  36.0\%&  &   $6.35\times 10^{-110}$ &36.6\%  &  &  $2.18\times 10^{-112}$  \\ \hline
			\textbf{Image2caption} &8.1\%  &  &  &10.1\%  &  &  & 10.7\%&  &  \\ \hline
		\end{tabular}
	}
\end{table*}

\section{Experiments on Comparison}
\label{sec:ec}

After the optimization of the method is handled by leveraging the pilot experiments described above, the model is then compared with an existing image caption generator and a poem generator with a large scale experiment.

\subsection{Experiment Setup}

\textbf{Test Image Data }
While we use randomly crawled images as testing queries in the previous experiment, we report our results on the existing Microsoft COCO dataset for image describing tasks\cite{chen2015microsoft} \cite{lin2014microsoft}. We use 270 images (recognized as view) in the validation set of COCO for the competitive analysis experiment.\\
\textbf{Assessors }
To obtain rich feedback without user bias, 22 assessors from variety career fields are chosen, including: 1) 8 female users and 14 male users, 2) 13 users with bachelor degree and 1 user with master or higher degree, 3) 11 users prefer traditional poetry, 10 users prefer modern poetry and 1 user prefers neither.\\   
\textbf{Baselines }
We compare with two state-of-the-art methods.
\begin{itemize}
\item\textbf{Image2caption} 
Since we are trying to describe an image, a first-class performance approach of image captioning \cite{fang2015captions} is chosen as a baseline according to the leader board on the COCO dataset. While the approach generates a one-line English sentence from a given image query, we translate the caption into Chinese by human effort and separate it into multiple lines (e.g., usually two lines) with appropriate Chinese grammar.
\item\textbf{CTRIP}
At the final stage of our research, we noticed that CTRIP released a traditional poetry generation application on 2017 \cite{Ctrip}. While we cannot find the corresponding publication, since it has a similar goal, we choose their released service as the second baseline in the comparison experiment. 47 image queries fail in their poetry generation process, though the specific reason why are unknown due to lack of technical detail. To maintain fairness, we only evaluate the performance on other 223 images for all three methods.
\end{itemize}
\begin{figure}[!h]
	\centering
	\includegraphics[width=0.9\columnwidth]{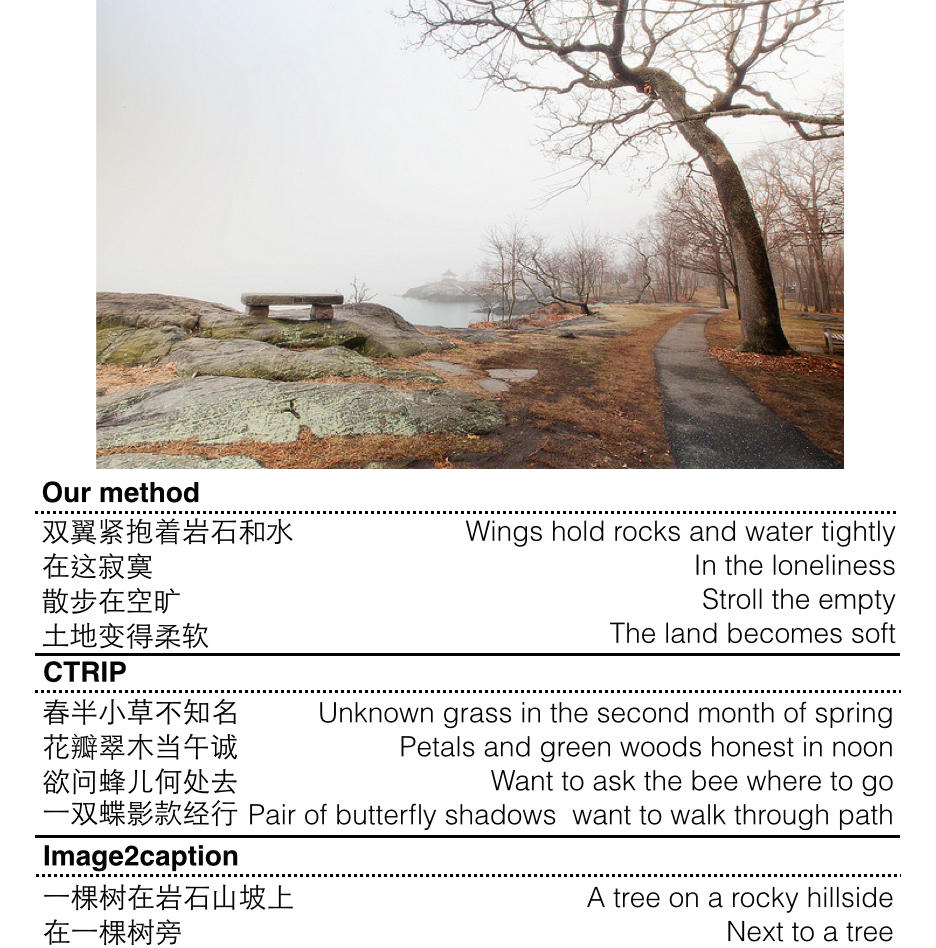}
	\caption{The example poems generated by two baselines method and our proposed method.     
		\label{fig:compete}
	}	
\end{figure} 

\textbf{Evaluation Metrics }
Beside the 1-to-5 overall rating for each approach, we also ask the assessors to give votes to the best methods in terms of five different criteria: relevant, fluent, imaginative, touching, and impressive. We provide multi-choice check boxes for each aspect and ask them to vote. We accumulate votes from all assessors for all images and then divide the number by $number \ of \ image \ queries *number \ of \  assessors$. The percentage of votes obtained are shown in Table~\ref{ex4}.

\subsection{Experiment Results}
As Table~\ref{ex4} indicates, both CTRIP and our method significantly outperform the image2caption method in all but Relevant measures, which indicates that describing images with poetry is more attractive. Our methods is significantly better than the two baselines in terms of being \textit{imaginative}, \textit{touching} and \textit{impressive}. This indicates that our proposed method is more effective to bring objects and feelings into modern poetry and thus our generated results are more imaginative and touching. With full understanding of the meaning, attention paid to modern poems can be also sustained longer. Thus our method is the best in being impressive. Along with the advantages, our method sacrifices some relevance compared to image2caption and fluency compared to CTRIP. 

We show three generated poems from one image in Figure~\ref{fig:compete}. The image2caption describes images with a straightforward way and thus most relevant. The poem generated by CTRIP is fluent and enjoyable in rhythms, but it brings the ``pair of butterfly'' that do not have natural association with the image in semantic or sentiment. Our poem is related to image, but bring some feelings like ``loneliness'' into the poem. When a user reads ``Stroll the empty / The land becomes soft'', he/she may get soft and hopeful too. This example shows that our modern poems are more easier to stimulates interesting ideas and generates emotional resonance.

In a summary, while the ``relevant'' part is dominated by image2caption and CTRIP is stronger in ``fluent'', our method significantly outperforms both baselines in other three aspects related to being artistic. The best overall score is also achieved by our proposed method.

\section{Conclusion and Future Work}
\label{sec:cf}
This paper introduces an innovative idea for generating an artistic poem from an image. Our best proposed solution includes a hierarchical model from sentence to poem for poem generation, a deep learning based keyword extraction and a statistical based keyword expansion. A large scale user study indicates that our generated modern poems earn much more favor than the generated captions. The reasons are that our generated poems are more imaginative, touching and impressive. Although the generated traditional poems are more fluent than our modern poems, the method we proposed significantly outperform them in all other three aspects.  

This study is our first attempt to generate poetry from image. In this study, while some connection between image and poetry is leverage, more can be done. For example, when we expand keywords, it is possible to verify if the additional keywords have some connection with the image content. It may be desirable that they carry some sort of connection with the image. In the same way, the poem generation step is separated from image, once the keywords have been extracted. It may also be desirable that generated sentences correspond better to the image. We will investigate approaches for this in the future.
\fontsize{9.0pt}{10.0pt} \selectfont
\bibliographystyle{aaai} 
\bibliography{reference}

\begin{thebibliography}{}

\bibitem[\protect\citeauthoryear{Bernardi \bgroup et al\mbox.\egroup
  }{2016}]{bernardi2016automatic}
Bernardi, R.; Cakici, R.; Elliott, D.; Erdem, A.; Erdem, E.; Ikizler-Cinbis,
  N.; Keller, F.; Muscat, A.; and Plank, B.
\newblock 2016.
\newblock Automatic description generation from images: A survey of models,
  datasets, and evaluation measures.
\newblock {\em J. Artif. Intell. Res.(JAIR)} 55:409--442.

\bibitem[\protect\citeauthoryear{Borth \bgroup et al\mbox.\egroup }{2013}]{VSO}
Borth, D.; Ji, R.; Chen, T.; Breuel, T.; and Chang, S.-F.
\newblock 2013.
\newblock Large-scale visual sentiment ontology and detectors using adjective
  noun pairs.
\newblock In {\em ACM Multimedia},  223--232.

\bibitem[\protect\citeauthoryear{Chen \bgroup et al\mbox.\egroup
  }{2015}]{chen2015microsoft}
Chen, X.; Fang, H.; Lin, T.-Y.; Vedantam, R.; Gupta, S.; Doll{\'a}r, P.; and
  Zitnick, C.~L.
\newblock 2015.
\newblock Microsoft coco captions: Data collection and evaluation server.
\newblock {\em arXiv preprint arXiv:1504.00325}.

\bibitem[\protect\citeauthoryear{Ctrip}{2017}]{Ctrip}
Ctrip.
\newblock 2017.
\newblock Ctrip Xiao Shi Ji.
\newblock http://pages.c-ctrip.com/commerce/promote/201701/other/xsj/index.html.

\bibitem[\protect\citeauthoryear{Devlin \bgroup et al\mbox.\egroup
  }{2015}]{devlin2015language}
Devlin, J.; Cheng, H.; Fang, H.; Gupta, S.; Deng, L.; He, X.; Zweig, G.; and
  Mitchell, M.
\newblock 2015.
\newblock Language models for image captioning: The quirks and what works.
\newblock {\em arXiv preprint arXiv:1505.01809}.

\bibitem[\protect\citeauthoryear{Donahue \bgroup et al\mbox.\egroup
  }{2015}]{donahue2015long}
Donahue, J.; Anne~Hendricks, L.; Guadarrama, S.; Rohrbach, M.; Venugopalan, S.;
  Saenko, K.; and Darrell, T.
\newblock 2015.
\newblock Long-term recurrent convolutional networks for visual recognition and
  description.
\newblock In {\em Proceedings of the IEEE conference on computer vision and
  pattern recognition},  2625--2634.

\bibitem[\protect\citeauthoryear{Fang \bgroup et al\mbox.\egroup
  }{2015}]{fang2015captions}
Fang, H.; Gupta, S.; Iandola, F.; Srivastava, R.~K.; Deng, L.; Doll{\'a}r, P.;
  Gao, J.; He, X.; Mitchell, M.; Platt, J.~C.; et~al.
\newblock 2015.
\newblock From captions to visual concepts and back.
\newblock In {\em Proceedings of the IEEE Conference on Computer Vision and
  Pattern Recognition},  1473--1482.

\bibitem[\protect\citeauthoryear{Fu \bgroup et al\mbox.\egroup
  }{2015}]{TaggingTransferDeep}
Fu, J.; Mei, T.; Yang, K.; Lu, H.; and Rui, Y.
\newblock 2015.
\newblock Tagging personal photos with transfer deep learning.
\newblock In {\em WWW},  344--354.

\bibitem[\protect\citeauthoryear{Ghazvininejad \bgroup et al\mbox.\egroup
  }{2016}]{ghazvininejad2016generating}
Ghazvininejad, M.; Shi, X.; Choi, Y.; and Knight, K.
\newblock 2016.
\newblock Generating topical poetry.
\newblock In {\em Proceedings of the 2016 Conference on Empirical Methods in
  Natural Language Processing},  1183--1191.

\bibitem[\protect\citeauthoryear{He, Zhou, and Jiang}{2012}]{he2012generating}
He, J.; Zhou, M.; and Jiang, L.
\newblock 2012.
\newblock Generating chinese classical poems with statistical machine
  translation models.
\newblock In {\em AAAI}.

\bibitem[\protect\citeauthoryear{Jiang and Zhou}{2008}]{jiang2008generating}
Jiang, L., and Zhou, M.
\newblock 2008.
\newblock Generating chinese couplets using a statistical mt approach.
\newblock In {\em Proceedings of the 22nd International Conference on
  Computational Linguistics-Volume 1},  377--384.
\newblock Association for Computational Linguistics.

\bibitem[\protect\citeauthoryear{Karpathy and
  Li}{2015}]{KarpathyL15}
Karpathy, A., and Li, F.
\newblock 2015.
\newblock Deep visual-semantic alignments for generating image descriptions.
\newblock In {\em {IEEE} Conference on Computer Vision and Pattern Recognition,
  {CVPR} 2015, Boston, MA, USA, June 7-12, 2015},  3128--3137.

\bibitem[\protect\citeauthoryear{Kingma and Ba}{2014}]{kingma2014adam}
Kingma, D., and Ba, J.
\newblock 2014.
\newblock Adam: A method for stochastic optimization.
\newblock {\em arXiv preprint arXiv:1412.6980}.

\bibitem[\protect\citeauthoryear{Krizhevsky, Sutskever, and
  Hinton}{2012}]{imagenet}
Krizhevsky, A.; Sutskever, I.; and Hinton, G.~E.
\newblock 2012.
\newblock Imagenet classification with deep convolutional neural networks.
\newblock In {\em NIPS},  1097--1105.

\bibitem[\protect\citeauthoryear{Lin \bgroup et al\mbox.\egroup
  }{2014}]{lin2014microsoft}
Lin, T.-Y.; Maire, M.; Belongie, S.; Hays, J.; Perona, P.; Ramanan, D.;
  Doll{\'a}r, P.; and Zitnick, C.~L.
\newblock 2014.
\newblock Microsoft coco: Common objects in context.
\newblock In {\em European Conference on Computer Vision},  740--755.
\newblock Springer.

\bibitem[\protect\citeauthoryear{Liu \bgroup et al\mbox.\egroup
  }{2016}]{liu2016not}
Liu, C.-W.; Lowe, R.; Serban, I.~V.; Noseworthy, M.; Charlin, L.; and Pineau,
  J.
\newblock 2016.
\newblock How not to evaluate your dialogue system: An empirical study of
  unsupervised evaluation metrics for dialogue response generation.
\newblock {\em arXiv preprint arXiv:1603.08023}.

\bibitem[\protect\citeauthoryear{Manurung, Ritchie, and
  Thompson}{2012}]{manurung2012using}
Manurung, R.; Ritchie, G.; and Thompson, H.
\newblock 2012.
\newblock Using genetic algorithms to create meaningful poetic text.
\newblock {\em Journal of Experimental \& Theoretical Artificial Intelligence}
  24(1):43--64.

\bibitem[\protect\citeauthoryear{Manurung}{2004}]{manurung2004evolutionary}
Manurung, H.
\newblock 2004.
\newblock An evolutionary algorithm approach to poetry generation.

\bibitem[\protect\citeauthoryear{Mikolov \bgroup et al\mbox.\egroup
  }{2010}]{mikolov2010recurrent}
Mikolov, T.; Karafi{\'a}t, M.; Burget, L.; Cernock{\`y}, J.; and Khudanpur, S.
\newblock 2010.
\newblock Recurrent neural network based language model.
\newblock In {\em Interspeech}, volume~2, ~3.

\bibitem[\protect\citeauthoryear{Netzer \bgroup et al\mbox.\egroup
  }{2009}]{netzer2009gaiku}
Netzer, Y.; Gabay, D.; Goldberg, Y.; and Elhadad, M.
\newblock 2009.
\newblock Gaiku: Generating haiku with word associations norms.
\newblock In {\em Proceedings of the Workshop on Computational Approaches to
  Linguistic Creativity},  32--39.
\newblock Association for Computational Linguistics.

\bibitem[\protect\citeauthoryear{Oliveira}{2012}]{oliveira2012poetryme}
Oliveira, H.~G.
\newblock 2012.
\newblock Poetryme: a versatile platform for poetry generation.
\newblock {\em Computational Creativity, Concept Invention, and General
  Intelligence} 1:21.

\bibitem[\protect\citeauthoryear{Patterson \bgroup et al\mbox.\egroup
  }{2014}]{patterson2014sun}
Patterson, G.; Xu, C.; Su, H.; and Hays, J.
\newblock 2014.
\newblock The sun attribute database: Beyond categories for deeper scene
  understanding.
\newblock {\em International Journal of Computer Vision} 108(1-2):59--81.

\bibitem[\protect\citeauthoryear{Schwarz, Berg, and
  Lensch}{2016}]{schwarz2016auto}
Schwarz, K.; Berg, T.~L.; and Lensch, H.~P.
\newblock 2016.
\newblock Auto-illustrating poems and songs with style.
\newblock In {\em Asian Conference on Computer Vision},  87--103.
\newblock Springer.

\bibitem[\protect\citeauthoryear{Shum, He, and Li}{2018}]{xiaoice}
Shum, H.-Y.; He, X.; and Li, D.
\newblock 2018.
\newblock From eliza to xiaoice: Challenges and opportunities with social
  chatbots.

\bibitem[\protect\citeauthoryear{Socher \bgroup et al\mbox.\egroup
  }{2014}]{Socher2014GroundedCS}
Socher, R.; Karpathy, A.; Le, Q.~V.; Manning, C.~D.; and Ng, A.~Y.
\newblock 2014.
\newblock Grounded compositional semantics for finding and describing images
  with sentences.
\newblock {\em TACL} 2:207--218.

\bibitem[\protect\citeauthoryear{Soto \bgroup et al\mbox.\egroup
  }{2015}]{DBLP:journals/aimatters/SotoKKM15}
Soto, A.~J.; Kiros, R.; Keselj, V.; and Milios, E.~E.
\newblock 2015.
\newblock Machine learning meets visualization for extracting insights from
  text data.
\newblock {\em {AI} Matters} 2(2):15--17.

\bibitem[\protect\citeauthoryear{Szegedy \bgroup et al\mbox.\egroup
  }{2015}]{googleNet}
Szegedy, C.; Liu, W.; Jia, Y.; Sermanet, P.; Reed, S.~E.; Anguelov, D.; Erhan,
  D.; Vanhoucke, V.; and Rabinovich, A.
\newblock 2015.
\newblock Going deeper with convolutions.
\newblock In {\em CVPR},  1--9.

\bibitem[\protect\citeauthoryear{Tosa, Obara, and Minoh}{2008}]{tosa2008hitch}
Tosa, N.; Obara, H.; and Minoh, M.
\newblock 2008.
\newblock Hitch haiku: An interactive supporting system for composing haiku
  poem.
\newblock In {\em International Conference on Entertainment Computing},
  209--216.
\newblock Springer.

\bibitem[\protect\citeauthoryear{Wang \bgroup et al\mbox.\egroup
  }{2016a}]{wang2016chinese}
Wang, Q.; Luo, T.; Wang, D.; and Xing, C.
\newblock 2016.
\newblock Chinese song iambics generation with neural attention-based model.
\newblock {\em arXiv preprint arXiv:1604.06274}.

\bibitem[\protect\citeauthoryear{Wang \bgroup et al\mbox.\egroup}{2016b}]{wang2016chinese2}
Wang, Z.; He, W.; Wu, H.; Wu, H.; Li, W.; Wang, H.; and Chen, E.
\newblock 2016.
\newblock Chinese poetry generation with planning based neural network.
\newblock {\em arXiv preprint arXiv:1610.09889}.

\bibitem[\protect\citeauthoryear{Wu, Tosa, and Nakatsu}{2009}]{wu2009new}
Wu, X.; Tosa, N.; and Nakatsu, R.
\newblock 2009.
\newblock New hitch haiku: An interactive renku poem composition supporting
  tool applied for sightseeing navigation system.
\newblock In {\em International Conference on Entertainment Computing},
  191--196.
\newblock Springer.

\bibitem[\protect\citeauthoryear{Yan \bgroup et al\mbox.\egroup
  }{2013}]{yan2013poet}
Yan, R.; Jiang, H.; Lapata, M.; Lin, S.-D.; Lv, X.; and Li, X.
\newblock 2013.
\newblock i, poet: Automatic chinese poetry composition through a generative
  summarization framework under constrained optimization.
\newblock In {\em IJCAI}.

\bibitem[\protect\citeauthoryear{Yan}{2016}]{yan2016poet}
Yan, R.
\newblock 2016.
\newblock i, poet: Automatic poetry composition through recurrent neural
  networks with iterative polishing schema.
\newblock IJCAI.

\bibitem[\protect\citeauthoryear{Yi, Li, and Sun}{2016}]{yi2016generating}
Yi, X.; Li, R.; and Sun, M.
\newblock 2016.
\newblock Generating chinese classical poems with rnn encoder-decoder.
\newblock {\em arXiv preprint arXiv:1604.01537}.

\bibitem[\protect\citeauthoryear{Zhang and Lapata}{2014}]{zhang2014chinese}
Zhang, X., and Lapata, M.
\newblock 2014.
\newblock Chinese poetry generation with recurrent neural networks.
\newblock In {\em EMNLP},  670--680.


\end{thebibliography}

\end{document}